\PassOptionsToPackage{dvipsnames}{xcolor}
\documentclass[11pt]{article}
\usepackage{acl}
\usepackage{times}
\usepackage{latexsym}
\usepackage{multirow}
\usepackage{amsmath,amsfonts}
\usepackage{amssymb,bm}
\usepackage[dvipsnames]{xcolor}
\usepackage[T1]{fontenc}
\usepackage[utf8]{inputenc}
\usepackage{microtype}
\usepackage{graphicx}
\usepackage{booktabs}
\usepackage{pifont}
\usepackage{array}
\usepackage{url}
\usepackage[caption=false,font=normalsize,labelfont=sf,textfont=sf]{subfig}
\usepackage{pgfplots}
\pgfplotsset{compat=1.18}
\usetikzlibrary{plotmarks}

\title{Conditioned in Training, Dispensable at Inference:\\
Typological Conditioning for Multilingual Speech Translation}
\author{
  Yu Pan$^{1}$\thanks{Corresponding author.} \quad
  Tianjiao Zhang$^{2}$ \quad
  Yang Hou$^{3}$ \quad
  Xiongfei Wu$^{4}$ \quad
  Liang Zhang$^{5}$ \\
  \textbf{Yves Le Traon}$^{4}$ \quad
  \textbf{Lei Ma}$^{6}$ \quad
  \textbf{Jianjun Zhao}$^{1}$ \\
  $^{1}$Kyushu University \quad
  $^{2}$Independent Researcher \quad
  $^{3}$National Institute of Informatics \\
  $^{4}$University of Luxembourg \quad
  $^{5}$Donghua University \quad
  $^{6}$The University of Tokyo \\
  \texttt{<panyu.ztj@gmail.com>}
}

\begin{document}
\maketitle

\begin{abstract}
Multilingual speech-to-speech translation frontends built on speech large language models typically condition on language identity, as a learned embedding or a prompted language name. Such a signal, however, is learned separately for each language and transfers nothing between typologically related languages. In this paper, we present TypoCast, a frontend that conditions on the typological profile of the source language rather than on its label alone. A typology-informed hierarchical language encoding represents each language by its morphological type, constituent order and family, so that supervision is shared among related languages. A dual-route mixture-of-experts adapter routes inputs through a shared typological route together with a per-language residual, and a typology-aware prompt conveys the same profile to the decoder. Extensive experiments on CVSS-C demonstrate that TypoCast surpasses state-of-the-art speech-to-speech translation systems on French, German and Spanish to English, averaged over the three directions, and outperforms the identity-conditioned S2ST-Omni on every language of a seven-language inventory, with only $228$M trainable parameters. Ablation studies further reveal that the conditioning signal acts during training and is dispensable at inference: a model trained without it is consistently worse, whereas removing it from a trained model changes BLEU by less than $0.2$, and expert usage is consistent with the signal shaping how adapter capacity is divided among languages.
\end{abstract}

\section{Introduction}
\label{sec:intro}

Multilingual speech-to-speech translation (S2ST) increasingly pairs a speech-to-text translation
(S2TT) frontend built on a speech large language model (LLM) with a separate synthesis
backend~\cite{fang2024can,zhang2024streamspeech,pan2025s2st,deng-etal-2025-simuls2s,chen2026llas2s}.
Because one frontend serves many source languages, existing systems inform it of the
source language by conditioning on language identity, a \emph{language-as-label} paradigm in which each source
language is an isolated symbol realized as a language token, adapter or
expert~\cite{johnson2017google,pfeiffer2020mad,ansell-etal-2023-unifying}.

A label, however, specifies \emph{which} language the model hears but conveys nothing
about \emph{what} that language is like.
Languages share regularities of morphology, constituent order and genealogy that bear
directly on aligning speech to source content and reordering the target
clause~\cite{comrie1989language,littell2017uriel,ponti2019modeling,oncevay2020bridging}.
A flat identity embedding exposes none of these regularities~\cite{rajaee-monz-2024-analyzing},
which leads to \textbf{identity isolation}: each embedding is learned only from its own
language's data, so Japanese and Turkish, both agglutinative and verb-final, have, by
construction, embeddings no more related than those of Japanese and French
(Figure~\ref{fig:overview}a). Such isolation forgoes whatever related languages could
share, a cost that is most acute in corpora as imbalanced as CVSS-C~\cite{jia2022cvss},
where French has $185$ times as many training utterances as Japanese.

\begin{figure*}[t]
\centering
\begin{tikzpicture}[
  lang/.style={anchor=east, font=\small, inner sep=1pt},
  cnt/.style={anchor=west, font=\scriptsize, text=black!55, inner sep=1pt},
  chip/.style={draw=black!40, rounded corners=1.5pt, minimum width=0.66cm,
    minimum height=0.34cm, inner sep=1pt, font=\scriptsize},
  idc/.style={chip, fill=black!8, minimum width=0.9cm},
  mor/.style={chip, fill=NavyBlue!15},
  ord/.style={chip, fill=Orange!25},
  fam/.style={chip, fill=Plum!20},
  resc/.style={chip, fill=ForestGreen!22},
  head/.style={font=\scriptsize, text=black!60},
  ptitle/.style={anchor=west, font=\small\bfseries},
  note/.style={anchor=north west, font=\scriptsize, align=left, text=black!80}]
\node[ptitle] at (-0.9,0.8) {(a) Identity conditioning};
\node[head] at (0.75,0.33) {identity};
\node[head, anchor=west] at (1.25,0.33) {utts.};
\foreach \l/\n/\y in {fr/207k/0, de/128k/-0.42, es/79k/-0.84, zh/7.1k/-1.26, tr/4.0k/-1.68, cy/1.2k/-2.10, ja/1.1k/-2.52} {
  \node[lang] at (0,\y) {\l};
  \node[idc] at (0.75,\y) {$\mathbf{e}_{\mathrm{\l}}$};
  \node[cnt] at (1.25,\y) {\n};
}
\node[note] at (-0.9,-2.85) {One free vector per language:\\ related languages share nothing.};
\begin{scope}[xshift=4.75cm]
\node[ptitle] at (-0.9,0.8) {(b) Shared attribute code};
\node[head] at (0.45,0.33) {morph.};
\node[head] at (1.17,0.33) {order};
\node[head] at (1.89,0.33) {family};
\foreach \l/\m/\o/\f/\y in {fr/fus/svo/Rom/0, de/fus.c/v2/Ger/-0.42, es/fus/svo/Rom/-0.84, zh/iso/svo/Sin/-1.26, tr/agg/sov/Tur/-1.68, cy/fus/vso/Cel/-2.10, ja/agg/sov/Jap/-2.52} {
  \node[lang] at (0,\y) {\l};
  \node[mor] (b\l) at (0.45,\y) {\m};
  \node[ord] at (1.17,\y) {\o};
  \node[fam] at (1.89,\y) {\f};
}
\draw[BrickRed, line width=1pt, rounded corners=2pt]
  ([shift={(-1.5pt,-1.5pt)}]bcy.south west) rectangle ([shift={(1.5pt,1.5pt)}]bcy.north east);
\node[note] at (-0.9,-2.85) {Related languages share class embeddings, but\\ \textcolor{BrickRed}{fus} is fitted to fr and es, with $231\times$ the data of cy.};
\end{scope}
\begin{scope}[xshift=10.1cm]
\node[ptitle] at (-0.9,0.8) {(c) TypoCast: dual-route conditioning};
\node[head] at (1.17,0.33) {shared typological route};
\node[head] at (2.97,0.33) {residual};
\foreach \l/\m/\o/\f/\y in {fr/fus/svo/Rom/0, de/fus.c/v2/Ger/-0.42, es/fus/svo/Rom/-0.84, zh/iso/svo/Sin/-1.26, tr/agg/sov/Tur/-1.68, cy/fus/vso/Cel/-2.10, ja/agg/sov/Jap/-2.52} {
  \node[lang] at (0,\y) {\l};
  \node[mor] at (0.45,\y) {\m};
  \node[ord] at (1.17,\y) {\o};
  \node[fam] at (1.89,\y) {\f};
  \node[font=\scriptsize] at (2.43,\y) {$+$};
  \node[resc] at (2.97,\y) {$\mathbf{b}_{\mathrm{\l}}$};
}
\node[note] at (-0.9,-2.85) {Shared class embeddings plus a zero-initialized\\ per-language residual intended to let a language\\ depart from its class as its own data warrant.};
\end{scope}
\end{tikzpicture}
\caption{Three ways to condition a multilingual frontend, shown for the seven languages
of Setting~B. (a) Identity conditioning isolates languages: Japanese and Turkish share
nothing despite identical morphology and order. (b) A purely shared attribute code allows
sharing, but each class embedding is fitted mostly to its best-resourced members. (c)
TypoCast pairs a shared typological route with a per-language residual. Counts are
training utterances.}
\label{fig:overview}
\end{figure*}

Typological attributes offer a natural remedy, since an attribute embedding receives
supervision from every language that shares it. They raise, however, a second problem, which we term
\textbf{majority-profile capture}: a shared class embedding is dominated by the
best-resourced members of its class, so a low-resource member inherits a prior fitted to
languages that may differ from it on every attribute the class does not constrain
(Figure~\ref{fig:overview}b).
Both problems concern the \emph{content} of the conditioning signal. A more fundamental
question, to our knowledge not previously posed for speech-LLM frontends, is \emph{when}
its benefit arises: at
inference, from the information it supplies to a trained model, or during training, from
how it shapes that model. The two accounts make distinct predictions. If the signal acts
at inference, removing it from a trained model should be costly. If it acts during
training, removal should cost nothing, the content should matter only through the
partition it induces among the languages, and no correction applied at inference can undo
that partition.

In this paper, we propose \textbf{TypoCast}, a frontend that conditions each
language-dependent module on the typological profile of the source language rather than
on its label alone, while retaining the backbone and training recipe of
S2ST-Omni (Figure~\ref{fig:overview}c). The profile enters the frontend at three sites
(Figure~\ref{fig:typocast}, Appendix~\ref{app:arch}).
To share supervision among related languages, a \emph{typology-informed hierarchical
language encoding} (TI-HLE) represents each language by the embeddings of its
morphological type, constituent order and family. To allocate adapter capacity by profile,
a \emph{dual-route mixture-of-experts} (MoE) adapter routes inputs through a shared
typological route together with a zero-initialized per-language residual, intended to let
a minority member depart from its class as its own data warrant. To convey the same
description to the decoder, a \emph{typology-aware prompt} replaces the language name
with a description of the profile.
Our main contributions are fourfold:
\begin{itemize}\setlength{\itemsep}{1pt}\setlength{\parskip}{0pt}
\item We show that language conditioning in a speech-LLM translation frontend acts while
the model is trained and is dispensable at inference: removing it from a trained model
changes the seven-language average by less than $0.2$ BLEU, whereas training without it
costs $1.50$ BLEU on average and up to $2.94$ at the router, with the same asymmetry at
the prompt. Expert usage supports an account in which the signal shapes how adapter
capacity is divided among languages.
\item We present TypoCast, to our knowledge the first frontend to condition a speech
translation LLM on typological attributes rather than language identity. On seven languages
with $207$k to $1.1$k training utterances, it reaches $29.77$ BLEU and $27.88$ ASR-BLEU
averaged over languages, surpassing S2ST-Omni trained with the same recipe by $2.78$ BLEU
on average and on every language.
\item We characterize majority-profile capture, show that it is measurable in the trained
router, and confirm two implications of the allocation account: the attributes act
jointly, so that the completeness and correctness of the description matter (a deranged
profile is worth no more than none), and a routing-level correction does not undo an
allocation, even at a competitive scale.
\item We demonstrate that the proposed TypoCast surpasses state-of-the-art (SOTA) S2ST systems on CVSS-C French, German and Spanish on ASR-BLEU, BLEU, BLASER~2.0 and
COMET averaged over the three directions, while updating only $228$M of $4.9$B parameters
on two A6000 GPUs.
\end{itemize}

\section{Related Work}
\label{sec:related}

\begin{table}[t]
\centering\small
\setlength{\tabcolsep}{4pt}
\caption{Languages of the two settings: the three above the rule form Setting~A, where
the attributes coincide (French and Spanish against German); all seven form Setting~B,
where no two attributes induce the same partition. Welsh is the sole member of two of its
classes and a minority member of the third.}
\label{tab:typology}
\renewcommand{\arraystretch}{0.8}
\begin{tabular}{lrlll}
\toprule
Lang. & Utts. & Morphology & Order & Family \\
\midrule
fr & 207{,}364 & fusional      & svo & Romance \\
de & 127{,}822 & fus.\ compound & v2  & Germanic \\
es & \phantom{0}79{,}012 & fusional      & svo & Romance \\
\midrule
zh & \phantom{00}7{,}085 & isolating     & svo & Sinitic \\
tr & \phantom{00}3{,}966 & agglutinative & sov & Turkic \\
cy & \phantom{00}1{,}241 & fusional      & vso & Celtic \\
ja & \phantom{00}1{,}119 & agglutinative & sov & Japonic \\
\bottomrule
\end{tabular}
\end{table}

\paragraph{Speech translation with speech LLMs.} S2ST has
been pursued as cascades~\cite{kikui2003creating}, end-to-end models that map source
speech to target speech or discrete
units~\cite{jia19_interspeech,jia2022translatotron,lee2022textless,inaguma2023unity,barrault2023seamlessm4t,fang2023daspeech},
and compositional models~\cite{fang2024can,zhang2024streamspeech,zheng2025rosettaspeech}
that pair a translation frontend, most recently a speech
LLM~\cite{deng-etal-2025-simuls2s,pan2025s2st,chen2026llas2s}, with a synthesis backend.
Our baseline, S2ST-Omni~\cite{pan2025s2st}, bridges a frozen Whisper encoder and an LLM
through a hybrid adapter and conditions a dual CTC objective and the prompt on language
identity; TypoCast retains its backbone and recipe, replaces the dense adapter with a
mixture of experts, and changes the conditioning variable.

\paragraph{Language conditioning in multilingual models.} Multilingual models receive the
source language as prepended tokens~\cite{johnson2017google}, parameter-generating or
modulating embeddings~\cite{platanios2018contextual,pan2025s2st}, language-specific
adapters~\cite{pfeiffer2020mad,ansell-etal-2023-unifying} or language-based expert
routing~\cite{kudugunta2021beyond,li-etal-2026-lcma}. These mechanisms differ in where
the condition enters but agree on its content: each language receives its own parameters,
unrelated to any other's. Adapters generated from typological vectors are the exception,
explored for dependency parsing~\cite{ustun2020udapter} but not for speech, where
attribute-structured signals have so far served emotion
recognition~\cite{pan2024gemoclap,pan2024gmptl}. TypoCast replaces that content with
typological attributes and, beyond content, asks when the condition acts at all.

\paragraph{Linguistic typology in NLP.} Typological knowledge, surveyed
by~\citet{ponti2019modeling} and encoded in URIEL~\cite{littell2017uriel}, has guided
multilingual parsing~\cite{ammar2016many}, transfer-language
selection~\cite{lin2019choosing} and machine translation~\cite{oncevay2020bridging}, and
is reflected in learned language vectors~\cite{ostling2017continuous}. In speech, it has
so far determined the grouping of shared connectors~\cite{zhang-etal-2026-language} or
experts~\cite{lee2026molge} for multilingual recognition. TypoCast is, to our knowledge,
the first to use several typological attributes as a soft conditioning signal for speech
translation, and in doing so characterizes majority-profile capture, a class-level form of
negative interference~\cite{wang2020negative}.

\section{TypoCast}
\label{sec:method}


\subsection{Preliminaries and Problem Analysis}
\label{sec:diagnosis}

\paragraph{Frontend.} TypoCast builds on the S2TT frontend of S2ST-Omni~\cite{pan2025s2st}:
a frozen Whisper-Large-v3 encoder~\cite{radford2023robust}, a hybrid convolution and
self-attention adapter into the embedding space of a Qwen3-4B decoder~\cite{yang2025qwen3},
and the decoder itself, adapted with LoRA~\cite{hu2022lora}. The source language enters at
three points (Figure~\ref{fig:typocast}, Appendix~\ref{app:arch}): the alignment of adapter
features to source content, supervised by a language-aware dual CTC
objective~\cite{graves2006connectionist,pan2025s2st}, the allocation of adapter capacity,
and the instruction given to the decoder. S2ST-Omni
conditions the first and third on language identity, uses a dense adapter, and identifies
the source language with the Whisper decoder, whereas all our experiments supply it
(Appendix~\ref{app:details}).

\paragraph{Typological profiles.} We characterize each training language $\ell\in\mathcal{L}$
by a profile $a(\ell)=\big(a_{\mathrm{m}}(\ell),a_{\mathrm{o}}(\ell),a_{\mathrm{f}}(\ell)\big)$
comprising its morphological type, constituent-order class and language family, drawn from
standard typological resources~\cite{littell2017uriel,comrie1989language}
(Table~\ref{tab:typology}).\footnote{Values follow the WALS categories in URIEL; German
is coded as its own morphological class (fusional with productive compounding) and as
verb-second, since a coarser coding would merge it with Romance on morphology.} Each value
of an attribute defines a class of languages.

\paragraph{Identity isolation.} Identity conditioning assigns each language a free vector
$\mathbf{e}_\ell$. The gradient that updates $\mathbf{e}_{\mathrm{ja}}$ conveys nothing to
$\mathbf{e}_{\mathrm{tr}}$, however similar the two languages may be, so whatever the
conditioning pathway learns about a profile must be relearned separately for each of its
members.

\paragraph{Majority-profile capture.} An attribute-keyed code instead shares one
embedding $\mathbf{u}_c$ among the members of class $c$. Since each member contributes
gradient in proportion to its data volume $n_\ell$, $\mathbf{u}_c$ is, to a first
approximation, a data-weighted average of the values that would serve each member alone:
\begin{equation}
\mathbf{u}_c \;\approx\; \sum_{\ell \in c} \frac{n_\ell}{\sum_{\ell' \in c} n_{\ell'}}\, \mathbf{u}_c^{(\ell)},
\label{eq:capture}
\end{equation}
where $\mathbf{u}_c^{(\ell)}$ denotes the value optimal for $\ell$ alone. A dominant member thus pulls
$\mathbf{u}_c$ toward its own optimum, and a minority member receives a prior fitted to
languages that agree with it only on the attribute defining $c$.
Equation~\ref{eq:capture} therefore implies that sharing benefits least not the languages
with the least data as such, but the minority members of classes whose other members
differ from them (tested in \S\ref{sec:boundary}).

\subsection{Typology-Informed Hierarchical Language Encoding}
\label{sec:tihle}

To overcome identity isolation, TI-HLE represents a language by the embeddings of its
attribute classes rather than by a single identity vector, with a per-language channel
$\mathbf{e}_\ell$ retaining what the attributes do not capture. The four channels are
concatenated and fused as follows:
\begin{equation}
\mathbf{z}_\ell = \mathrm{GELU}\big(\mathrm{LN}(\mathbf{W}[\mathbf{u}_{a_{\mathrm{m}}(\ell)};\mathbf{u}_{a_{\mathrm{o}}(\ell)};\mathbf{u}_{a_{\mathrm{f}}(\ell)};\mathbf{e}_\ell])\big),
\label{eq:tihle}
\end{equation}
where $[\cdot\,;\cdot]$ denotes concatenation and bias terms are omitted. Each attribute
embedding is updated by every language in its class, so supervision is shared among
typologically related languages, and because the encoding is composed of named channels,
any single attribute can be ablated without altering the others.

\paragraph{Conditioning the alignment branch.} TI-HLE replaces the identity embedding in
the language-aware CTC branch of S2ST-Omni, which aligns adapter features to source
content. A FiLM generator~\cite{perez2018film} maps
$\mathbf{z}_\ell$ to a bounded scale and shift that modulate each adapter feature
$\mathbf{h}_t$ through a frame-level gate, formulated as
\begin{equation}
\begin{aligned}
{[\boldsymbol{\gamma};\boldsymbol{\beta}]} &= \tanh\big(f(\mathbf{z}_\ell)\big),\\
g_t &= \sigma\big(f_g([\mathbf{h}_t;\mathbf{z}_\ell])/\tau\big),\\
\widetilde{\mathbf{h}}_t &= (\mathbf{1}+g_t\boldsymbol{\gamma})\odot\mathbf{h}_t+g_t\boldsymbol{\beta},
\end{aligned}
\label{eq:film}
\end{equation}
where $f$ and $f_g$ are small feed-forward networks, $\tau$ is a learnable temperature,
and the gate bias is initialized so that the modulation is initially weak. The modulated
features feed the two CTC heads only, whereas the decoder receives the unmodulated adapter
output, so this conditioning site is active only during training.

\subsection{Dual-Route Mixture-of-Experts Adapter}
\label{sec:dualroute}

As a candidate remedy for majority-profile capture, which \S\ref{sec:correction} tests, we
replace the feed-forward sublayers
of the adapter's attention blocks with a mixture of experts~\cite{shazeer2017outrageous}
whose router receives two language-dependent terms. For an adapter feature
$\mathbf{x}_t$ from language $\ell$, the router computes logits over $N$ experts as
\begin{equation}
\mathbf{s}_t \;=\; \underbrace{\mathbf{W}_{\mathrm{r}}\mathbf{x}_t}_{\text{content}}
\;+\; \underbrace{\mathbf{W}_{\mathrm{s}}\,\mathbf{z}^{\mathrm{typ}}_{\ell}}_{\text{shared}}
\;+\; \underbrace{\mathbf{b}_{\ell}}_{\text{residual}},
\label{eq:dual}
\end{equation}
where $\mathbf{z}^{\mathrm{typ}}_{\ell}$ is a separate instance of
Equation~\ref{eq:tihle} over the three attribute channels only and
$\mathbf{b}_\ell\in\mathbb{R}^N$ is a per-language bias.
Top-2 gating selects two of $N{=}4$ routed experts, and an always-active shared expert
processes every input. We refer to the shared and residual terms collectively as the
router's \emph{language terms}. The shared route depends on $\ell$ only through $a(\ell)$, so languages with the
same profile share expert preferences, whereas the residual route is per-language and is
intended to let a minority member depart from the fit of its class. Both $\mathbf{W}_{\mathrm{s}}$ and the residual biases are zero-initialized, so
training starts from content-only routing and any typological structure in expert usage
is learned rather than imposed.

\subsection{Typology-Aware Prompt and Training}
\label{sec:prompt}

\paragraph{Typology-aware prompt.} To convey the typological description to the decoder
as well, the prompt replaces the language name with a one-sentence description of the
profile and three hand-written translation guidelines derived from it, appended to a
shared instruction (Appendix~\ref{app:prompts}). Languages with the same profile receive an
identical prompt, and no prompt contains utterance-specific information.

\paragraph{Training and inference.} Training follows the two stages of S2ST-Omni:
Stage~1 trains the adapter, the conditioning modules and the CTC branches with the decoder
frozen, and Stage~2 resumes from the best Stage~1 checkpoint and adds LoRA, with a
load-balancing loss on expert usage throughout. At inference the CTC branch and its FiLM
modulation are discarded, so the router's two language terms and the prompt are the only
inputs through which the given source language reaches the model, and
\S\ref{sec:where} removes each of them in turn.

\section{Experimental Setup}
\label{sec:setup}

\paragraph{Datasets.} We conduct experiments on CVSS-C~\cite{jia2022cvss}, built on
CoVoST~2~\cite{wang21s_interspeech}, under two settings (Table~\ref{tab:typology}).
\textbf{Setting~A} covers French, German and Spanish to English, the configuration
adopted in prior S2ST work, and serves to compare TypoCast with existing S2ST systems. In
this setting the three attributes partition the languages identically (French and Spanish
versus German), so their effects cannot be separated. \textbf{Setting~B} covers all seven
languages, chosen among the CVSS-C source languages so that no two attributes induce the
same grouping, and hosts the ablations. Morphology and constituent order are crossed, whereas family refines both and, with six classes over seven languages, is close to a per-language code.

\begin{table*}[t]
\centering
\caption{Setting~A results on CVSS-C. Bold: best; ``-'': not reported. Greyed rows are reference points, not highlighted: ground-truth bound, text cascade, model of a different scale (Appendix~\ref{app:metrics}). $^\dagger$Single many-to-one model. TypoCast is given
the source language (Appendix~\ref{app:details}).}
\label{tab:main}
\resizebox{\textwidth}{!}{
\begin{tabular}{lcccccccc}
\hline
\multirow{2}{*}{Model}
& \multicolumn{2}{c}{Fr$\to$En}
& \multicolumn{2}{c}{De$\to$En}
& \multicolumn{2}{c}{Es$\to$En}
& \multicolumn{2}{c}{Average} \\
\cline{2-9}
& BLEU & ASR-BLEU & BLEU & ASR-BLEU & BLEU & ASR-BLEU & BLEU & ASR-BLEU \\
\hline
\textcolor{black!40}{Ground Truth}
& \textcolor{black!40}{-} & \textcolor{black!40}{84.52}
& \textcolor{black!40}{-} & \textcolor{black!40}{75.53}
& \textcolor{black!40}{-} & \textcolor{black!40}{88.54}
& \textcolor{black!40}{-} & \textcolor{black!40}{82.86} \\
\hline
\multicolumn{9}{@{}l}{\emph{End-to-end}} \\
Translatotron~2~\cite{jia2022translatotron}
& 28.82 & 26.07 & 18.66 & 16.91 & 25.82 & 22.93 & 24.43 & 21.97 \\
StreamSpeech~\cite{zhang2024streamspeech}
& 32.60 & 28.45 & 23.36 & 20.93 & 30.35 & 27.25 & 28.77 & 25.54 \\
SLM-S2ST$^\dagger$~\cite{hu2025slm}
& - & 35.93 & - & 35.54 & - & 39.67 & - & 37.05 \\
\textcolor{black!40}{Qwen3-Omni-30B~\cite{qwen3omni2025}}
& \textcolor{black!40}{43.65} & \textcolor{black!40}{39.56}
& \textcolor{black!40}{43.48} & \textcolor{black!40}{39.02}
& \textcolor{black!40}{46.12} & \textcolor{black!40}{41.31}
& \textcolor{black!40}{44.41} & \textcolor{black!40}{39.96} \\
\hline
\multicolumn{9}{@{}l}{\emph{Cascaded}} \\
\textcolor{black!40}{Whisper--Qwen3 (text)}
& \textcolor{black!40}{35.15} & \textcolor{black!40}{-}
& \textcolor{black!40}{36.07} & \textcolor{black!40}{-}
& \textcolor{black!40}{38.39} & \textcolor{black!40}{-}
& \textcolor{black!40}{36.54} & \textcolor{black!40}{-} \\
\hline
\multicolumn{9}{@{}l}{\emph{Compositional}} \\
ComSpeech~\cite{fang2024can}
& 30.72 & 28.15 & 19.41 & 18.16 & 26.51 & 24.80 & 25.55 & 23.70 \\
RosettaSpeech$^\dagger$~\cite{zheng2025rosettaspeech}
& 33.11 & 32.16 & 23.22 & 21.54 & 30.92 & 29.35 & 29.08 & 27.68 \\
S2ST-Omni$^\dagger$~\cite{pan2025s2st}
& 35.83 & 33.20 & 33.34 & 31.25 & 37.85 & 35.90 & 35.67 & 33.45 \\
LLaS2S-2.3B~\cite{chen2026llas2s}
& - & 35.46 & - & 33.54 & - & 38.08 & - & 35.69 \\
\hline
\textbf{TypoCast}$^\dagger$
& \textbf{41.51} & \textbf{37.98} & \textbf{38.42} & \textbf{35.56} & \textbf{43.37} & \textbf{40.43} & \textbf{41.10} & \textbf{37.99} \\
\hline
\end{tabular}}
\end{table*}

\paragraph{Baselines.} In Setting~A, we compare TypoCast with seven existing S2ST
systems, both end-to-end~\cite{jia2022translatotron,zhang2024streamspeech,hu2025slm} and
compositional~\cite{fang2024can,zheng2025rosettaspeech,chen2026llas2s,pan2025s2st}, and
include as a reference point a text cascade that translates Whisper-Large-v3 transcripts
with Qwen3-4B. Each Setting~B variant shares with the full model the encoder, decoder,
adapter parameter budget, optimization schedule, batch size, number of training steps in
each stage, checkpoint-selection rule and decoding procedure, and differs from it in
exactly one component.

\paragraph{Evaluation.} For the text output, we report BLEU~\cite{post-2018-call} and
COMET~\cite{rei2022comet}; for the speech output, ASR-BLEU computed on transcriptions
from the wav2vec~2.0 recognizer used in prior CVSS-C work, and
BLASER~2.0~\cite{dale2024blaser2}. Speech is synthesized from a single reference speech
with the zero-shot synthesizer VoxCPM2~\cite{zhou2026voxcpm2}; Appendix~\ref{app:tts}
shows that the choice of synthesizer does not affect the ordering of systems.
Metric configurations and hyperparameters are given in
Appendix~\ref{app:details}.

\section{Main Results}
\label{sec:results}

\subsection{Comparison with Existing S2ST Systems}
\label{sec:setA}

\begin{table}[t]
\centering
\caption{Speech-level BLASER~2.0 (reference-based) and text-level COMET
(\texttt{wmt22-comet-da}, 0--100) on CVSS-C under Setting~A. Bold: best; ``-'': not
reported; greyed rows are reference points.}
\label{tab:blaser}
\small
\setlength{\tabcolsep}{3pt}
\resizebox{\columnwidth}{!}{%
\begin{tabular}{lcccccccc}
\hline
\multirow{2}{*}{Model} & \multicolumn{4}{c}{BLASER~2.0} & \multicolumn{4}{c}{COMET} \\
\cline{2-9}
& Fr & De & Es & Avg. & Fr & De & Es & Avg. \\
\hline
\textcolor{black!40}{Qwen3-Omni-30B~\cite{qwen3omni2025}} & \textcolor{black!40}{4.18} & \textcolor{black!40}{4.14} & \textcolor{black!40}{4.32} & \textcolor{black!40}{4.21} & \textcolor{black!40}{-} & \textcolor{black!40}{-} & \textcolor{black!40}{-} & \textcolor{black!40}{-} \\
\textcolor{black!40}{Whisper\,+\,Qwen3 (cascade)} & \textcolor{black!40}{-} & \textcolor{black!40}{-} & \textcolor{black!40}{-} & \textcolor{black!40}{-} & \textcolor{black!40}{80.20} & \textcolor{black!40}{83.36} & \textcolor{black!40}{83.79} & \textcolor{black!40}{82.45} \\
\hline
Translatotron~2~\cite{jia2022translatotron} & 3.18 & 2.90 & 3.26 & 3.11 & - & - & - & - \\
ComSpeech~\cite{fang2024can} & 3.19 & 2.93 & 3.28 & 3.13 & 70.13 & 57.96 & 66.96 & 65.02 \\
StreamSpeech~\cite{zhang2024streamspeech} & 3.20 & 3.00 & 3.31 & 3.17 & 76.66 & 65.51 & 74.80 & 72.32 \\
RosettaSpeech~\cite{zheng2025rosettaspeech} & 4.04 & 4.08 & 4.17 & 4.10 & 78.97 & 79.65 & 82.05 & 80.22 \\
LLaS2S-2.3B~\cite{chen2026llas2s} & 4.12 & 3.93 & 4.22 & 4.09 & - & - & - & - \\
S2ST-Omni~\cite{pan2025s2st} & 4.12 & \textbf{4.10} & 4.21 & 4.14 & 81.94 & 80.73 & 83.39 & 82.02 \\
\hline
\textbf{TypoCast} & \textbf{4.18} & 4.09 & \textbf{4.30} & \textbf{4.19} & \textbf{84.01} & \textbf{83.18} & \textbf{86.18} & \textbf{84.46} \\
\hline
\end{tabular}}
\end{table}

\paragraph{TypoCast leads on both speech-level metrics.} Table~\ref{tab:main} compares
TypoCast with existing S2ST systems in Setting~A, with systems ranked by ASR-BLEU since
the task produces speech. TypoCast attains an average of $37.99$, $0.94$ above SLM-S2ST,
the strongest existing system on this metric, and $2.30$ above LLaS2S-2.3B, and leads in
all three directions. The choice of synthesizer shifts this average by at most $0.21$
(Appendix~\ref{app:tts}). Its $41.10$ text BLEU exceeds the best reported S2ST figure,
$35.67$, by $5.43$, and its BLASER~2.0 of $4.19$ leads S2ST-Omni's $4.14$
(Table~\ref{tab:blaser}), trailing only in German, by $0.01$.

\paragraph{Not an artefact of surface overlap.} COMET, which also scores against the
source sentence and so does not reward lexical overlap with the reference alone
(Table~\ref{tab:blaser}), preserves the ordering of the existing systems and ranks
TypoCast highest at $84.46$, ahead of S2ST-Omni by $2.07$, $2.45$ and $2.79$ on the three
directions. German again yields the narrowest separation: on ASR-BLEU, BLEU and
BLASER~2.0 TypoCast's margin over the strongest S2ST system is smallest for German. It is also the only Setting~A source that is the sole member of each of its
attribute classes, so the shared route can transfer nothing to it
(\S\ref{sec:correction}).

\paragraph{Scale and paradigm.} Qwen3-Omni-30B~\cite{qwen3omni2025}, a general-purpose
model of a different scale, leads TypoCast by $3.31$ text BLEU but by only $1.97$ ASR-BLEU
and $0.02$ BLASER~2.0 ($0.04$ in the reference-free variant): a large text-BLEU gap need
not carry over to the delivered speech (Appendix~\ref{app:metrics}).

\subsection{The Seven-Language Inventory}
\label{sec:setB}

Setting~B (Table~\ref{tab:perlang}) carries every ablation in this paper. The full
model attains $29.77$ BLEU, whereas S2ST-Omni re-implemented under the same recipe
attains $26.99$; TypoCast leads it on all seven languages, by $2.78$ on average and by
$0.57$ on Welsh, and the variant of \S\ref{sec:ablation} that replaces TI-HLE with an
identity embedding reaches only $27.88$ BLEU. The margin over S2ST-Omni is not confined
to languages with typological neighbours: German, the sole member of each of its classes,
gains $3.00$, as much as French and Spanish, indicating that the typological code changes
what every language's conditioning pathway learns and not only what related languages
share. Because the unweighted mean gives Welsh and Japanese, $2.8\%$ of the test set, two
sevenths of its weight, we also report the size-weighted mean, $37.17$ ($36.95$ pooled
over all $49{,}379$ utterances). Synthesizing and transcribing the outputs lowers the mean to $27.88$ ASR-BLEU
($34.55$ size-weighted) without changing the ordering of the languages, so the text-level
measurements are not an artefact of omitting the speech channel
(Appendix~\ref{app:setbspeech}).

\section{Analysis}
\label{sec:analysis}

\subsection{Conditioning Acts at Training Time}
\label{sec:where}

The language terms could act at inference, by selecting experts per input, or during
training, by shaping what the experts learn. To separate the two, we remove the terms from
the trained seven-language model at inference, decoding the full test sets of all seven
languages, and compare the outcome with training without them (Table~\ref{tab:ablation}).

\begin{table}[t]
\centering\small
\setlength{\tabcolsep}{5pt}
\caption{Ablations under Setting~B, average BLEU over 7 languages; $\Delta_{\mathrm{cy}}$ is the change on Welsh. The greyed row is S2ST-Omni re-implemented under Setting~B.}
\label{tab:ablation}
\renewcommand{\arraystretch}{0.95}
\resizebox{\columnwidth}{!}{%
\begin{tabular}{lrrr}
\toprule
Model & Avg BLEU & $\Delta$ & $\Delta_{\mathrm{cy}}$ \\
\midrule
\textcolor{black!40}{S2ST-Omni (same recipe)} & \textcolor{black!40}{26.99} & \textcolor{black!40}{$-2.78$} & \textcolor{black!40}{$-0.57$} \\
\textbf{TypoCast (full)} & \textbf{29.77} & -- & -- \\
\midrule
\multicolumn{4}{l}{\emph{Capacity vs.\ conditioning in the adapter}} \\
w/o residual route & 28.34 & $-1.43$ & $+0.48$ \\
w/o language terms (content-only) & 28.27 & $-1.50$ & $-0.19$ \\
w/o MoE (dense, matched params) & 28.52 & $-1.25$ & $+0.59$ \\
\midrule
\multicolumn{4}{l}{\emph{Attribute channels}} \\
\quad w/o morphology & 28.19 & $-1.58$ & $+0.24$ \\
\quad w/o family & 28.53 & $-1.24$ & $+0.27$ \\
\quad w/o reordering & 28.34 & $-1.43$ & $-0.77$ \\
\midrule
\multicolumn{4}{l}{\emph{Conditioning sites}} \\
w/o TI-HLE (per-language emb.) & 27.88 & $-1.89$ & $+0.47$ \\
w/o TA-Prompt (language name) & 28.67 & $-1.10$ & $+0.79$ \\
\midrule
\multicolumn{4}{l}{\emph{Profile content}} \\
deranged profiles (placebo) & 28.28 & $-1.49$ & $+0.55$ \\
\bottomrule
\end{tabular}}
\end{table}

\paragraph{Inference-time removal does not change BLEU.} The removal changes $11\%$ of the
$49{,}379$ hypotheses, mostly by rewording (Appendix~\ref{app:removal}), yet shifts the
seven-language mean by $0.04$ BLEU and no language by more than $0.15$ (Turkish $-0.15$,
Japanese $-0.14$, the other five within $0.05$), whereas training without the terms costs
$1.50$ on average and $2.94$, $1.10$ and $0.19$ on Turkish, Japanese and Welsh
(Table~\ref{tab:perlang}). The terms are dispensable rather than
ignored: assigning French the German profile in the Setting~A model costs $0.53$ BLEU,
whereas Spanish, the Romance minority with a residual of its own, loses $0.15$
(Appendix~\ref{app:capture}). The router still
responds to the signal, but what it carries is redundant with the content term once the
experts are trained.

\paragraph{Expert usage supports an allocation account.} The terms thus act during
training rather than by selecting experts at test time. Removing them at inference shifts
the expert distribution of each of Turkish, Japanese and Welsh by at most $0.02$ in total
variation (Appendix~\ref{app:experts}), whereas the model trained without them routes the
same languages differently ($0.06$ to $0.10$ with experts matched) and separates them
less. Nor is the shared term negligible, averaging about half the norm of the content term in
the first expert layer (Appendix~\ref{app:capture}).

\paragraph{The prompt shows the same asymmetry.} Replacing the typological clause with the
language name at inference shifts the seven-language mean by $0.07$ BLEU and no language by
more than $0.74$ (Turkish; the other six within $0.26$), whereas a model trained
throughout with the language-name prompt loses $1.10$ on average and $2.88$ on Turkish
(Table~\ref{tab:perlang}). Removing both at inference, leaving only the shared instruction
(Appendix~\ref{app:prompts}) and hence no language information, costs $0.18$ on average
and at most $0.66$ (Turkish), with $0.52$ on German and $0.44$ on Japanese. At both sites, then,
the signal is consumed while the model is trained. On the allocation account, two
implications follow and are tested next. First, the content of the signal matters only
through the partition it induces, so the attributes should act jointly
(\S\ref{sec:ablation}). Second, a term acting only on routing cannot repair an allocation
after the fact (\S\ref{sec:correction}).

\subsection{The Description Is Consumed as a Whole}
\label{sec:ablation}

Table~\ref{tab:ablation} ablates one component of the Setting~B model at a time under
the matched recipe of \S\ref{sec:setup}.

\paragraph{Identity in place of typology costs most.} Replacing TI-HLE with an identity
embedding of nearly the same size at both sites is the costliest change in the table,
$1.89$ BLEU ($29.77$ to $27.88$), more than the $1.50$ lost under content-only routing,
which removes the router's language terms but keeps the typological encoding in the
alignment branch. A signal that
partitions the languages into singletons is thus of less use than one that partitions them
by attribute, and of no more use than none (the identity variant also changes the
alignment branch), as the allocation account implies: a signal matters through the
partition it induces, and the identity partition shares nothing. Content-only routing
loses $1.50$ with experts, encoding and prompt all retained, and a dense adapter of matched
parameter count loses $1.25$, so the experts contribute through how their capacity is
allocated rather than through the capacity itself. Every removal degrades the six
languages other than Welsh, which declines only under content-only routing ($0.19$, \S\ref{sec:correction})
and order removal. Removing the residual route alone costs $1.43$, nearly as much as
removing both terms, so the routes are not additive and we do not apportion the gain
between them. As the trained residual is negligible in the seven-language model (norm at
most $0.03$, Appendix~\ref{app:capture}), its contribution too is made during training
rather than at inference.

\paragraph{A wrong description is worth no more than none.} To separate the content of
the description from the presence of a shared code, we train a placebo with deranged
profiles: every language receives another language's complete profile and prompt (French
the Celtic one, German and Turkish the Romance one), preserving class sizes, sharing
structure and prompt templates while making every attribute wrong. It reaches $28.28$
BLEU, $1.49$ below the full model and within $0.01$ of content-only routing, and lowers
every language but Welsh (Table~\ref{tab:perlang}), Turkish most ($3.86$): a shared code
that groups the wrong languages adds nothing to content routing. As prompt and profile
are deranged together, the loss is not apportioned between wrong partition and wrong
guidelines. The full model's margin therefore derives from the partition the true
profiles induce (French with Spanish, Turkish with Japanese) together with the
descriptions that accompany it.
\paragraph{The channels behave as one signal.} The three channels are the
constituents of TI-HLE, so the reference against which to measure them is its replacement
by an identity embedding, $1.89$ BLEU. Removing any one channel costs most of that amount, morphology
$1.58$, constituent order $1.43$ and family $1.24$ (a lower bound, as family is removed at
the router only, Appendix~\ref{app:channels}), and the three losses sum to $4.25$, more
than twice the whole, whereas independently consumed channels would each cost about a
third and sum to it. Although the inventory induces three different partitions
(\S\ref{sec:setup}) and the per-language channel $\mathbf{e}_\ell$ of
Equation~\ref{eq:tihle}, retained in every channel ablation, could absorb a removed
attribute, the channels act as one factor: what the model consumes is the completeness of
the description rather than any single attribute, as the allocation account implies. A
smaller channel-specific component is examined in Appendix~\ref{app:channels}.

\subsection{A Routing Correction Cannot Undo the Allocation}
\label{sec:correction}

If capture is decided while experts are allocated, a per-language bias on the router,
which can only redirect inputs among experts already shaped by the pooled gradient, should
be unable to recover a minority member. Welsh is the one language on which this can be
tested.
\paragraph{Welsh lacks an informative typological neighbourhood.} Welsh is the only
verb-initial and the only Celtic language in the inventory, so its order and family
embeddings are learned from its own $1{,}241$ utterances, whereas its one shared class,
fusional morphology, also contains French and Spanish with $231$ times as many and is
fitted, by Equation~\ref{eq:capture}, mainly to these subject-verb-object languages rather
than to verb-initial Welsh. Welsh gains when the borrowed morphology class ($+0.24$) or its private family class ($+0.27$)
is removed and loses $0.77$ only without constituent order, the one attribute on which it
differs from every other language: that channel alone encodes something true of Welsh
that the task uses, the borrowed class harms it, and the family class carries nothing it
can use.

\paragraph{Capture is measurable in the router.} Projected onto the mean direction of the
shared term (Table~\ref{tab:capture}), the Romance profile, the best-resourced, aligns at
$+0.980$ and Welsh at $+0.954$ despite sharing only its morphology with Romance, whereas
Japanese ($-0.729$) and German ($-0.910$) point elsewhere. The alignment is not an
artefact of the shared channel (Appendix~\ref{app:capture}). Welsh's residual bias has a norm of $0.011$, about
one twentieth of its shared term, too small to correct it.

\paragraph{The residual route does not recover Welsh, and scaling it up does not help.}
Removing the residual route leaves Welsh unharmed ($+0.48$, Table~\ref{tab:ablation}), and
its small magnitude could be a parameterization artefact, so we re-parameterize it to pass
through the same normalization and non-linearity as the shared term, holding everything
else fixed. Welsh's residual grows thirteenfold, to more than half the shared term, yet
its BLEU changes by $-0.14$ and the seven-language average by $-1.15$. Since the change is
not neutral for the other languages, the comparison establishes only that a residual
thirteen times larger does not lift Welsh above the full model, and Welsh being the least
affected language is no evidence of a correction, as its dependence on the conditioning
is the smallest in the inventory (\S\ref{sec:boundary}). A residual of competitive scale thus fails as a negligible one does, at the expert budget
we use, as the allocation account implies: the experts available to Welsh were shaped by the pooled gradient of its class,
and no router bias can assign it an expert trained for it. A remedy, if one exists, must act on the allocation itself, by re-weighting the
gradient within a class or reserving capacity for minority members from the start.

\subsection{Where the Benefit Thins}
\label{sec:boundary}

If the benefit arises from training-time allocation, its size should depend on how
informative a language's classes are about it, not on how much data the language
contributes.
\paragraph{Data volume does not predict dependence.} Taking each language's mean absolute
change over the five component ablations as its dependence on the conditioning pathway
(Table~\ref{tab:perlang}), we find no monotone relation to the number of training
utterances: Turkish, with $3{,}966$ utterances, depends on the conditioning most ($2.79$
BLEU) and Welsh, with $1{,}241$, least ($0.50$), with the three highest-resource languages
between them (likewise with the channel removals included or on a relative scale).

\paragraph{A neighbourhood criterion.} Together with the Welsh case of
\S\ref{sec:correction}, the other six languages support a criterion: Japanese depends on the conditioning nearly three
times as much as Welsh despite fewer utterances, because it shares both morphology and
order with Turkish (Appendix~\ref{app:perlang}). A typological prior thus helps a language
most when its neighbourhood is informative:
enough data of its own, a peer with the same profile, or a shared class correct on the
attribute the task needs. The criterion is read from Table~\ref{tab:typology}
rather than from corpus size and argues against a pure capacity account, which would not
concentrate the thinnest benefit on the language whose class composition predicts it.

\section{Conclusion}
In this paper, we introduce TypoCast, a speech-LLM translation frontend that conditions the alignment branch, the mixture-of-experts router and the prompt on the typological profile of the source language rather than on its label, so that supervision is shared among typologically related languages. Extensive experiments on CVSS-C demonstrate that TypoCast surpasses SOTA S2ST systems on French, German and Spanish to English, averaged over the three directions, and outperforms the
identity-conditioned S2ST-Omni on every language of a seven-language inventory, with only $228$M trainable parameters. Our analysis further reveals that the conditioning signal acts during training and is dispensable at inference, with expert usage supporting an account in which it shapes how adapter capacity is divided among languages. Consistent with this account, the three attributes act jointly, and a routing-level correction does not rescue a language from majority-profile capture, whose remedy must therefore act on
the allocation itself. We hope these findings encourage a shift in the design of multilingual conditioning from richer per-language representations at inference toward how the signal partitions the training data.

\section*{Limitations}
\paragraph{Typological attributes.} We use three coarse, discrete attributes from existing
resources. They are interpretable but simplify linguistic reality, assigning German, for
example, a single order class although its order differs between main and subordinate
clauses. The typology-aware prompt pairs the profile with hand-written guidelines, and its
ablation replaces both with the language name, so the prompt-site gain is not attributed
to the typological description alone; likewise the placebo deranges profile and prompt
together. The adapter uses four routed experts throughout, so the failure of the routing
correction (\S\ref{sec:correction}) is established at that budget only.
\paragraph{Inventory and scope.} Our analysis covers seven languages with a single
captured member, so the neighbourhood criterion and the residual route are tested on one
case. No language outside the inventory is evaluated, so transfer to unseen typological
profiles is not addressed. The source language is supplied, not identified; Whisper-Large-v3 identifies Welsh
correctly in only $83\%$ of test utterances (Appendix~\ref{app:details}), although a
system given no source language at all loses only $0.18$ BLEU on average
(\S\ref{sec:where}).

\bibliography{main}

\appendix
\section{Model Architecture}
\label{app:arch}
\begin{figure*}[t]
\centering
    \includegraphics[height=8.2cm,width=!]{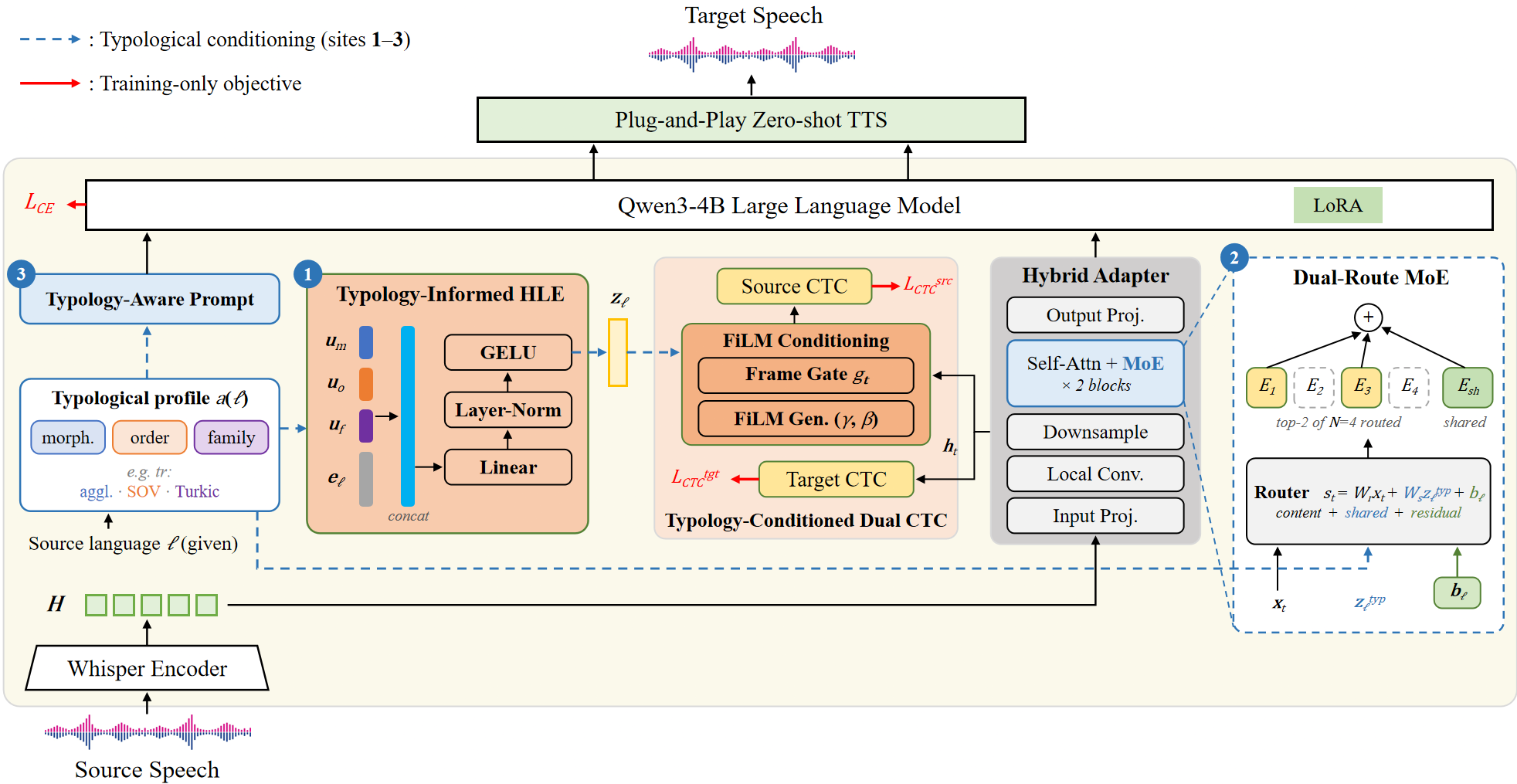}
    \caption{Overview of TypoCast. The source language is mapped to its typological profile $a(\ell)$, which conditions the frontend at three sites (dashed blue): (1)~TI-HLE, whose encoding $\mathbf{z}_\ell$ modulates adapter features through gated FiLM in the dual CTC branch; (2)~the dual-route MoE router of the hybrid adapter, with a shared typological term and a per-language residual $\mathbf{b}_\ell$; (3)~the typology-aware prompt. Red arrows mark training-only objectives.}
    \label{fig:typocast}
\end{figure*}

Figure~\ref{fig:typocast} depicts the TypoCast frontend with the three conditioning sites
of \S\ref{sec:method} and the training-only objectives.

\section{Implementation and Evaluation Details}
\label{app:details}
\paragraph{Implementation.} LoRA (rank $8$, scaling $32$, dropout $0.1$) is applied to the
query and value projections. Each of the adapter's two mixture-of-experts layers has four
routed experts and one shared expert with top-2 gating and a load-balancing weight of
$0.01$. The source and target CTC losses are weighted by $0.1$ and $0.2$. Learning rates
are $10^{-5}$ for the adapter, $5{\times}10^{-5}$ for LoRA, and $5{\times}10^{-5}$ then
$2{\times}10^{-5}$ for the CTC branches in the two stages. We train in bf16 on two
RTX~A6000 GPUs with an effective batch of $66$ utterances in Stage~1 and $64$ in Stage~2,
and select checkpoints by validation loss. All Setting~B variants share the data pipeline,
sampler, optimization and checkpoint-selection configuration and differ only in the
ablated component. The TypoCast model is trained for $150$k
Stage~1 and $150$k Stage~2 steps, including each ablation, and decodes with beam search of
width four. The CTC heads and the FiLM generator are training-only and add no inference
cost.
\paragraph{Source language.} All of our experiments supply the gold source language,
whereas the published S2ST-Omni figures in Table~\ref{tab:main} use the language
predicted by the Whisper decoder. Supplying it serves two purposes: a wrong profile is not
harmless (assigning French the German profile costs $0.53$ BLEU, \S\ref{sec:where}), so
identification errors would confound the ablations, and every Setting~B variant then
receives identical inputs. The two protocols coincide on the Setting~A languages, where
Whisper-Large-v3 identifies the source language of $99.2\%$, $99.1\%$ and $99.0\%$ of the
French, German and Spanish test utterances. On the remaining languages it reaches
$99.4\%$ (zh), $97.2\%$ (tr), $98.4\%$ (ja) and $83.0\%$ (cy), $98.8\%$ size-weighted
over the $49{,}379$ utterances, and $7\%$ of Welsh utterances are identified as English. A
deployed system would therefore behave as reported except on Welsh, the language the
inventory supports least (\S\ref{sec:correction}).
\paragraph{Metrics.} ASR-BLEU transcribes with wav2vec~2.0 Large (LV-60) with self-training,
the recognizer of prior CVSS-C evaluations. BLEU is computed with SacreBLEU, signature
\texttt{nrefs:1|case:mixed|\allowbreak eff:no|tok:13a|\allowbreak smooth:exp|\allowbreak version:2.3.1};
COMET uses \texttt{wmt22-comet-da}, scored against the source sentence and the reference;
BLASER~2.0 uses the \texttt{blaser-2.0-ref} and \texttt{blaser-2.0-qe} models with SONAR
speech and text encoders.

\section{Per-Language Ablation Results}
\label{app:perlang}
Table~\ref{tab:perlang} expands every row of Table~\ref{tab:ablation} by language.
Columns are ordered by training corpus size, from French ($207{,}364$ utterances) to
Japanese ($1{,}119$), so that the absence of a relation between corpus size and
dependence on the conditioning is visible across the final row
(\S\ref{sec:boundary}). Against S2ST-Omni under the same recipe, TypoCast's margin is
largest for Turkish ($+4.27$) and smallest for Welsh ($+0.57$), the two languages that the
neighbourhood criterion places at the opposite ends of the inventory, with the other five
between $+2.50$ and $+3.16$.

Read language by language, that row follows the neighbourhood criterion. Japanese has fewer
utterances than Welsh ($1{,}119$ vs.\ $1{,}241$) yet loses $1.44$ BLEU on average when a component of the conditioning pathway is ablated,
against $0.50$ for Welsh, because it shares both its morphology
and its constituent order with Turkish and the two depart from English order in the same
way; Turkish, which has that same neighbour, is the most dependent language in the
inventory. Chinese is, like Welsh, the sole member of two of its three classes, but the
class it does share, subject-verb-object order, is correct for it, and it loses $1.39$. The
three highest-resource languages have sufficient data of their own. Welsh is the only language that combines an uninformative shared class with too little
data to shape an allocation of its own.

\begin{table}[h]
\centering\small
\setlength{\tabcolsep}{3.5pt}
\caption{Per-language BLEU of S2ST-Omni re-implemented under the Setting~B recipe and of
TypoCast, and the change from TypoCast for every Setting~B ablation; columns are ordered
by decreasing corpus size. The last row gives the mean absolute
change over the five component ablations, the dependence measure of
\S\ref{sec:boundary}.}
\label{tab:perlang}
\resizebox{\columnwidth}{!}{%
\begin{tabular}{lrrrrrrrr}
\toprule
Model & fr & de & es & zh & tr & cy & ja & Mean \\
\midrule
S2ST-Omni (same recipe) & 36.93 & 35.09 & 39.26 & 16.12 & 29.35 & 12.05 & 20.14 & 26.99 \\
TypoCast (full) & 40.09 & 38.09 & 42.10 & 19.23 & 33.62 & 12.62 & 22.64 & 29.77 \\
\midrule
\multicolumn{9}{@{}l}{\emph{Capacity vs.\ conditioning in the adapter}} \\
$\Delta$ w/o residual route & $-1.55$ & $-1.70$ & $-1.28$ & $-1.13$ & $-2.76$ & $+0.48$ & $-2.09$ & $-1.43$ \\
$\Delta$ w/o language terms & $-1.59$ & $-1.82$ & $-1.13$ & $-1.76$ & $-2.94$ & $-0.19$ & $-1.10$ & $-1.50$ \\
$\Delta$ w/o MoE           & $-1.64$ & $-1.49$ & $-1.30$ & $-1.44$ & $-2.42$ & $+0.59$ & $-1.03$ & $-1.25$ \\
\midrule
\multicolumn{9}{@{}l}{\emph{Attribute channels}} \\
$\Delta$ w/o morphology & $-1.94$ & $-2.01$ & $-1.64$ & $-1.69$ & $-2.37$ & $+0.24$ & $-1.63$ & $-1.58$ \\
$\Delta$ w/o family     & $-1.67$ & $-1.60$ & $-1.23$ & $-1.29$ & $-1.96$ & $+0.27$ & $-1.22$ & $-1.24$ \\
$\Delta$ w/o reordering & $-1.25$ & $-1.42$ & $-1.11$ & $-3.18$ & $-1.59$ & $-0.77$ & $-0.66$ & $-1.43$ \\
\midrule
\multicolumn{9}{@{}l}{\emph{Conditioning sites}} \\
$\Delta$ w/o TI-HLE    & $-2.26$ & $-2.43$ & $-1.91$ & $-1.77$ & $-2.94$ & $+0.47$ & $-2.38$ & $-1.89$ \\
$\Delta$ w/o TA-Prompt & $-1.55$ & $-1.62$ & $-0.96$ & $-0.87$ & $-2.88$ & $+0.79$ & $-0.59$ & $-1.10$ \\
\midrule
\multicolumn{9}{@{}l}{\emph{Profile content}} \\
$\Delta$ deranged profiles & $-1.49$ & $-1.45$ & $-1.34$ & $-1.29$ & $-3.86$ & $+0.55$ & $-1.58$ & $-1.49$ \\
\midrule
Mean $|\Delta|$ & 1.72 & 1.81 & 1.32 & 1.39 & \textbf{2.79} & \textbf{0.50} & 1.44 & \\
\bottomrule
\end{tabular}%
}
\end{table}

\section{Setting~B at the Speech Level}
\label{app:setbspeech}
Table~\ref{tab:setbspeech} gives the per-language figures summarized in \S\ref{sec:setB}.
The text-level outputs of the full Setting~B model are synthesized with VoxCPM2 from the
same reference voice as Setting~A and transcribed with wav2vec~2.0, so the two columns
score the same hypotheses before and after the speech channel. The loss from text to
speech is largest for the three high-resource languages, up to $3.06$ for French, and
smallest for Welsh and Japanese, at $0.30$ and $0.64$. The ranking of the seven languages
is identical under the two measurements.
\begin{table}[t]
\centering\small
\setlength{\tabcolsep}{5pt}
\caption{TypoCast under Setting~B at the text and speech level. Speech is synthesized
with VoxCPM2 from the Setting~A reference voice; ASR-BLEU uses wav2vec~2.0
transcriptions.}

\label{tab:setbspeech}
\begin{tabular}{lrrr}
\toprule
Lang. & BLEU & ASR-BLEU & $\Delta$ \\
\midrule
fr & 40.09 & 37.03 & $-3.06$ \\
de & 38.09 & 35.47 & $-2.62$ \\
es & 42.10 & 39.48 & $-2.62$ \\
zh & 19.23 & 16.97 & $-2.26$ \\
tr & 33.62 & 31.86 & $-1.76$ \\
cy & 12.62 & 12.32 & $-0.30$ \\
ja & 22.64 & 22.00 & $-0.64$ \\
\midrule
\multicolumn{1}{l}{Mean} & 29.77 & 27.88 & $-1.89$ \\
\multicolumn{1}{l}{Weighted} & 37.17 & 34.55 & $-2.62$ \\
\bottomrule
\end{tabular}
\end{table}

\section{Typology-Aware Prompts}
\label{app:prompts}
Every prompt begins with the shared instruction \emph{You are an expert in speech
translation. Your task is to translate speech into natural English text.} followed by the
profile description and guidelines below; languages with the same profile (French and
Spanish) share a prompt verbatim, and the prompts are identical in the two settings. The
intervention of \S\ref{sec:where} that removes the typological prompt uses this shared
instruction alone.

\begin{small}
\paragraph{fusional / svo / Romance (fr, es).} The following speech is in a fusional
Romance language whose subject-verb-object (SVO) order broadly aligns with English. When
translating into English: 1.~Map rich verbal and nominal inflection (tense, mood, and
gender/number agreement) onto the appropriate English forms. 2.~Render idiomatic and
figurative expressions by their intended meaning rather than word by word. 3.~Recover
pro-dropped subjects and resolve clitics and function words from context.

\paragraph{fus.\ compound / v2 / Germanic (de).} The following speech is in a fusional
Germanic language with productive nominal compounding and frequent verb-second and
clause-final constructions. When translating into English: 1.~Decompose compound nouns
into their component meaning and render them as natural English noun phrases. 2.~Reorder
verb-final and clause-final structures, including separable verb prefixes and subordinate
clauses, into English subject-verb-object order. 3.~Resolve case marking to preserve the
intended argument structure.

\paragraph{agglutinative / sov / Japonic (ja).} The following speech is in an
agglutinative Japonic language with subject-object-verb (SOV) order, particle-marked
arguments, and frequent argument omission. When translating into English: 1.~Reorder
subject-object-verb structure into English subject-verb-object order. 2.~Use case and
topic particles to assign grammatical roles, and recover omitted subjects and objects from
context. 3.~Convey politeness and honorific distinctions through natural English register
without adding content.

\paragraph{agglutinative / sov / Turkic (tr).} The following speech is in an agglutinative
Turkic language with subject-object-verb (SOV) order, long suffix chains that carry case,
possession, tense, and negation, and pervasive subject omission. When translating into
English: 1.~Reorder subject-object-verb structure into English subject-verb-object order.
2.~Unpack suffix chains into the corresponding English words, and recover omitted subjects
from person marking on the verb. 3.~Render postpositions and case marking as the
appropriate English prepositions.

\paragraph{fusional / vso / Celtic (cy).} The following speech is in a fusional Celtic
language with verb-subject-object (VSO) order, initial consonant mutation, and
prepositions that inflect for person. When translating into English: 1.~Reorder
verb-initial structure into English subject-verb-object order. 2.~Recover the underlying
form of mutated word-initial consonants before interpreting a word. 3.~Resolve inflected
prepositions into the corresponding English preposition plus pronoun.

\paragraph{isolating / svo / Sinitic (zh).} The following speech is in an isolating
Sinitic language with subject-verb-object (SVO) order, no inflectional morphology, and
grammatical relations expressed by word order, aspect markers, and measure words. When
translating into English: 1.~Supply the tense, number, and definiteness that the source
leaves unmarked, inferring them from context. 2.~Interpret aspect and modal particles as
the corresponding English tense and mood rather than translating them as words. 3.~Recast
topic-comment constructions and serial verbs into English clause structure.
\end{small}

\section{Outputs under Inference-Time Removal}
\label{app:removal}
Table~\ref{tab:removal} shows hypotheses of the seven-language model that change when
the router's language terms are removed at inference (\S\ref{sec:where}). Of the
$49{,}379$ test utterances, $5{,}358$ (10.9\%) change, from $7.8\%$ for
Spanish to $21.2\%$ for Welsh and $23.3\%$ for Chinese. The changed hypotheses have a chrF
of $61$ to $83$ against the originals, lowest for Welsh and highest for Spanish, and the
changes are almost always lexical or ordering variants of equal adequacy, as in the
examples, which were selected among changed hypotheses whose sentence-level chrF against
the reference is equal before and after the removal.
\begin{table}[h]
\centering\small
\setlength{\tabcolsep}{4pt}
\caption{Hypotheses of the seven-language model before and after removing the router's
language terms at inference.}
\label{tab:removal}
\resizebox{\columnwidth}{!}{%
\begin{tabular}{@{}lp{0.86\columnwidth}@{}}
\toprule
\multicolumn{2}{@{}l}{\emph{fr}\quad Ref.: next phrap assigns the confirmed base to the consensus sequence} \\
full & then frappe assigns the confirmed base to the consensus sequence \\
w/o & frappe then assigns the confirmed base to the consensus sequence \\
\midrule
\multicolumn{2}{@{}l}{\emph{de}\quad Ref.: he chose hamburg as place of publication because of the city's relatively mild censorship} \\
full & he chose hamburg as the place of publication because censorship was relatively mild there \\
w/o & he chose hamburg as the place of publication because there censorship was relatively mild \\
\midrule
\multicolumn{2}{@{}l}{\emph{es}\quad Ref.: an impenetrable darkness cold and covered by shadows was created} \\
full & he created an impenetrable darkness cold and covered by shadows \\
w/o & it created an impenetrable darkness cold and covered by shadows \\
\midrule
\multicolumn{2}{@{}l}{\emph{zh}\quad Ref.: li ruijun born in gaotai gansu is a film director in mainland china} \\
full & li ruijun was born in gaotai gan su china and is a film director \\
w/o & li ruijun was born in gaotai gansu china and is a film director \\
\bottomrule
\end{tabular}}
\end{table}

\section{Routing Geometry of the Trained Model}
\label{app:capture}
Table~\ref{tab:capture} reports the quantities used in \S\ref{sec:correction}, read from
the first adapter layer of the trained Setting~B model. For each of the six profiles we
compute the shared-route term $\mathbf{W}_{\mathrm{s}}\mathbf{z}^{\mathrm{typ}}_\ell$ of
Equation~\ref{eq:dual}, its norm, and its cosine with the mean of the six terms, so that an
alignment near $+1$ marks a profile whose expert preference coincides with the dominant
direction of the shared route. The Romance profile, which contributes most of the training
data, defines that direction, and the Celtic profile of Welsh lies almost on it despite
sharing only its morphology with Romance, whereas the Germanic and Japonic profiles point
the other way. Because the mean direction is dominated by the largest terms, we also compute
each profile's cosine with the mean of the other five profiles: Celtic remains the most
aligned at $+0.825$, Romance itself falls to $-0.161$, and Germanic to $-0.941$. Pairwise,
the Welsh term has cosine $+0.928$ with the Romance term and $-0.979$ with the Germanic
term, whereas Turkish and Japanese, which share two of three channels, have cosine
$+0.272$, so the proximity of Welsh to Romance is not explained by the one channel they
share. In the second mixture-of-experts layer the shared terms of all six profiles collapse
onto one direction (pairwise cosines $\geq0.75$), so the profile-specific structure
resides in the first layer, which the table reports. The residual biases $\mathbf{b}_\ell$
are omitted because their norms lie between $0.003$ and $0.03$ across languages and layers,
$0.011$ for Welsh in the first layer against a shared-term norm of $0.244$. Across the six
profiles the shared term averages $0.48$ times the norm of the content term in the first
layer and $0.18$ times in the second. In the
Setting~A model the first-layer shared term has norm $0.712$ for the Romance profile and
$1.054$ for German, with residuals of $0.044$ (French), $0.090$ (Spanish) and $0.017$
(German). Its shared term averages $1.29$ times the norm of the content term in the first
layer and $0.11$ times in the second. French contributes $72\%$ of the Romance class's
training data, so by Equation~\ref{eq:capture} the shared Romance term is fitted mainly to
it and French needs no residual, whereas Spanish, the minority member, learns one twice as
large. This explains why assigning the German profile (Germanic, verb-second) at inference
changes $31\%$ of French hypotheses and costs $0.53$ BLEU but changes $19\%$ of Spanish
hypotheses and costs only $0.15$ (\S\ref{sec:where}).
\begin{table}[t]
\centering\small
\setlength{\tabcolsep}{4pt}
\caption{Majority-profile capture in the trained Setting~B router (first adapter layer). Alignment is the cosine between a profile's shared-route term and the mean direction of the six terms.}
\label{tab:capture}
\resizebox{\columnwidth}{!}{%
\begin{tabular}{llrr}
\toprule
Profile & Langs & Align. & $\lVert\text{shared}\rVert$ \\
\midrule
fusional / svo / Romance      & es, fr & $+0.980$ & 0.532 \\
\textbf{fusional / vso / Celtic} & \textbf{cy} & $\bm{+0.954}$ & 0.244 \\
isolating / svo / Sinitic     & zh & $+0.439$ & 0.219 \\
agglut. / sov / Turkic        & tr & $+0.150$ & 0.135 \\
agglut. / sov / Japonic       & ja & $-0.729$ & 0.117 \\
fus.\ comp.\ / v2 / Germanic  & de & $-0.910$ & 0.282 \\
\bottomrule
\end{tabular}}
\end{table}

\section{Expert Usage under the Interventions}
\label{app:experts}
Table~\ref{tab:experts} reports, for each of the two mixture-of-experts layers, the fraction
of adapter frames that each language routes to each of the four routed experts (top-1
choice) on the Turkish, Japanese and Welsh test sets, for the full seven-language model,
for the same model with the router's language terms removed at inference, and for the model
trained without them (content-only routing, Table~\ref{tab:ablation}). Removing the terms
at inference leaves every language's distribution almost unchanged (total variation at
most $0.02$ in either layer) and leaves the languages as separated as before (mean
pairwise distance $0.126$ and $0.120$ in the two layers, against $0.143$ and $0.121$ for
the full model). The experts of an independently trained model carry no correspondence to
those of the full model, so for the content-only model we match experts by the
permutation of the four that minimizes the mean discrepancy over the three languages,
the table listing each model's experts in its own order. Even under this most favourable
matching its distributions differ from the full model's by $0.06$ to $0.10$, and it
separates the languages less: the mean pairwise distance, which does not depend on the
matching, falls to $0.084$ in the first layer and $0.111$ in the second.
The language terms thus determined, during training, which experts each language came to
use. Once trained, the content term alone reproduces that assignment.
\begin{table}[h]
\centering\small
\setlength{\tabcolsep}{3.5pt}
\caption{Top-1 expert usage per language (fraction of frames routed to experts 1--4) in
the two mixture-of-experts layers.}
\label{tab:experts}
\resizebox{\columnwidth}{!}{%
\begin{tabular}{lrrrrrrrr}
\toprule
& \multicolumn{4}{c}{Layer 1} & \multicolumn{4}{c}{Layer 2} \\
\cmidrule(lr){2-5}\cmidrule(lr){6-9}
Language & E1 & E2 & E3 & E4 & E1 & E2 & E3 & E4 \\
\midrule
\multicolumn{9}{@{}l}{\emph{Full model}} \\
Turkish & 0.22 & 0.34 & 0.16 & 0.27 & 0.30 & 0.25 & 0.23 & 0.22 \\
Japanese & 0.18 & 0.28 & 0.13 & 0.41 & 0.48 & 0.14 & 0.16 & 0.22 \\
Welsh & 0.26 & 0.28 & 0.23 & 0.23 & 0.31 & 0.24 & 0.24 & 0.22 \\
\midrule
\multicolumn{9}{@{}l}{\emph{Full model, language terms removed at inference}} \\
Turkish & 0.22 & 0.33 & 0.17 & 0.28 & 0.28 & 0.27 & 0.23 & 0.22 \\
Japanese & 0.18 & 0.28 & 0.14 & 0.41 & 0.46 & 0.16 & 0.17 & 0.22 \\
Welsh & 0.24 & 0.28 & 0.24 & 0.24 & 0.29 & 0.24 & 0.24 & 0.22 \\
\midrule
\multicolumn{9}{@{}l}{\emph{Trained without language terms}} \\
Turkish & 0.20 & 0.20 & 0.29 & 0.31 & 0.34 & 0.21 & 0.24 & 0.20 \\
Japanese & 0.19 & 0.20 & 0.22 & 0.39 & 0.38 & 0.21 & 0.25 & 0.16 \\
Welsh & 0.26 & 0.23 & 0.20 & 0.32 & 0.24 & 0.33 & 0.22 & 0.21 \\
\bottomrule
\end{tabular}}
\end{table}

\section{Attribute Channels on a Relative Scale}
\label{app:channels}
Morphology and constituent order are removed from both sites TI-HLE serves. In the
alignment branch, family is not exposed separately from the per-language channel
$\mathbf{e}_\ell$ of Equation~\ref{eq:tihle}, so family is removed from the router only and
its figure in Table~\ref{tab:ablation} is a lower bound.
We therefore do not rely on the ordering of the three magnitudes.
The crossed inventory makes a sharper question testable than \S\ref{sec:ablation}
reports: whether a channel matters most to the languages whose behaviour it encodes.
Welsh, which improves when morphology or family is removed, is treated in
\S\ref{sec:correction}. For the other six languages we report each change both in BLEU
points and as a fraction of the language's own score, since their absolute quality spans
$19.2$ to $42.1$ BLEU and a fixed number of points is not the same loss for each of them.
Removing constituent order costs Chinese most, $3.18$ points or $16.5\%$ of its score,
against $0.66$ to $1.59$ points for the other five: subject-verb-object order is the one
class that Chinese shares with other languages, French and Spanish, and it is correct for
it, so removing the channel removes its only shared class, as the neighbourhood criterion
of \S\ref{sec:boundary} implies. The two verb-final languages lose $1.12$ points, or
$3.8\%$, against $1.26$ points, or $3.2\%$, for French, German and Spanish, so the order
channel is not consumed only by the languages whose order departs from English. Removing
family costs the four languages that are the sole members of their family $1.52$ points
against $1.45$ for the Romance pair, a separation that emerges only on the relative
scale, at $5.5\%$ against $3.5\%$. The two scalings agree on the conclusions of
\S\ref{sec:ablation}, and dependence on the conditioning remains unrelated to corpus size.

\section{Text versus Speech-Level Metrics}
\label{app:metrics}
We relate text-level and speech-level measurements through
Qwen3-Omni-30B~\cite{qwen3omni2025}, an end-to-end model of a different scale and paradigm
that Tables~\ref{tab:main} and~\ref{tab:blaser} include as a reference point. Its
advantage over TypoCast shrinks as evaluation moves from text to speech, from $3.31$ points
in text BLEU (its text output scored after the same normalization as the references) to
$1.97$ in ASR-BLEU and $0.02$ in reference-based BLASER~2.0 ($4.21$
against $4.19$), with $0.04$ in the reference-free variant ($4.24$ against $4.20$): a
large text-BLEU gap need not imply a comparable gap in the delivered speech. TypoCast
retains $92\%$ of its text BLEU in the speech channel and Qwen3-Omni $90\%$, which is
consistent with the compositional design handing the synthesizer its text verbatim,
whereas Qwen3-Omni generates speech separately from its text; the two systems differ in
scale, training data and synthesizer as well, so we draw no conclusion about the two
paradigms from this one pair.
We do not read the reference-based figure as parity, since BLASER~2.0 compresses in this
range, placing RosettaSpeech at $4.10$ despite a text BLEU of $29.08$, and Qwen3-Omni is
evaluated zero-shot whereas TypoCast is trained in domain with $228$M of its $4.9$B
parameters updated; Qwen3-Omni also generates text and speech independently, and our
speech-level figures depend on the synthesizer, though not in their ordering
(Appendix~\ref{app:tts}).

\section{Robustness to the Synthesis Backend}
\label{app:tts}

\begin{table}[h]
\centering\small
\setlength{\tabcolsep}{4pt}
\caption{Effect of the synthesis backend with the frontend's Setting~A outputs held
fixed: ASR-BLEU and reference-based BLASER~2.0 averaged over the three directions on the
full test sets. AR/NAR: autoregressive/non-autoregressive decoding.}
\label{tab:tts}
\resizebox{\columnwidth}{!}{%
\begin{tabular}{lllrr}
\toprule
Backend & Representation & Decoding & ASR-BLEU & BLASER \\
\midrule
VoxCPM2~\cite{zhou2026voxcpm2}      & continuous & AR  & 37.99 & 4.19 \\
Qwen3-TTS~\cite{qwen3tts2026}       & discrete   & AR  & 37.92 & 4.19 \\
IndexTTS-2~\cite{zhou2025indextts2} & discrete   & AR  & 37.85 & 4.20 \\
OmniVoice~\cite{omnivoice2026}      & discrete   & NAR & 37.78 & 4.18 \\
\bottomrule
\end{tabular}}
\end{table}

A compositional design makes every speech-level measurement depend on a synthesizer
outside the contribution being evaluated. Table~\ref{tab:tts} holds the frontend's
Setting~A outputs fixed and varies only the backend across systems that span both
representations of neural speech generation, continuous acoustic latents and discrete
codec~\cite{pan2024promptcodec} or semantic tokens, and both decoding regimes, autoregressive and
non-autoregressive: VoxCPM2 generates continuous latents autoregressively, Qwen3-TTS and
IndexTTS-2 generate discrete tokens autoregressively, and OmniVoice generates discrete
tokens non-autoregressively by masked diffusion; VoxCPM2 is the backend of Tables~\ref{tab:main}, \ref{tab:blaser} and~\ref{tab:setbspeech}. On the full test sets the four backends lie
within $0.21$ ASR-BLEU and $0.02$ BLASER~2.0 of one another, and the ordering of the
averages in Tables~\ref{tab:main} and~\ref{tab:blaser} is unchanged under every backend,
so the speech-level conclusions rest on neither a particular synthesizer nor a particular
family of them.

\end{document}